\documentclass[conference]{IEEEtran}

\usepackage{cite}
\usepackage{amsmath,amssymb,amsfonts}
\usepackage{algorithm, algpseudocode}
\usepackage{graphicx}
\usepackage{textcomp}
\usepackage{xcolor}
\usepackage{siunitx}

\def\BibTeX{{\rm B\kern-.05em{\sc i\kern-.025em b}\kern-.08em
    T\kern-.1667em\lower.7ex\hbox{E}\kern-.125emX}}

\newcommand{\minisection}[1]{\vspace{0.05in} \noindent {\bf #1} \ }
\usepackage[numbers]{natbib}

\usepackage{url}
\PassOptionsToPackage{hyphens}{url}\usepackage{hyperref}
\usepackage[most]{tcolorbox}

\begin{document}

\newcommand{\algName}{\text{RAID}}
\newcommand{\algNameLong}{\text{Reward-Adaptive Iterative Discovery}}

\title{\algNameLong: A Case Study on Automated Game Testing for NHL26
}

\author{\IEEEauthorblockN{Florian Fuchs\textsuperscript{*}, Jessy Gosselin-Grant, Boris Skuin, Michele Petteni,\\Alessandro Sestini, Joakim Bergdahl, Amir Baghi, Linus Gisslén\textsuperscript{*}}
\IEEEauthorblockN{\textit{Electronic Arts (EA)}\\
 \{ffuchs,lgisslen\}@ea.com \\
 \textsuperscript{*}Corresponding authors
}}

\maketitle

\begin{abstract}
Testing is a major effort for the gaming industry, requiring a significant part of development budget and people power.
We present a case study on a development version of the ice hockey game \textit{EA SPORTS NHL 26}, for which human playtesters test the goalie AI for behavioral exploits.
To reduce the effort of re-testing the goalie AI after every game or behavior modification in the development phase, we propose \algNameLong\ (\algName), a novel approach to automatically find exploits using an iterative Reinforcement Learning (RL) approach that trains a population of goal scoring agents.
While previous approaches can already successfully find exploits, RL algorithms tend to overfit to a single solution.
We introduce a simple extension on top of existing RL algorithms, such that they find multiple diverse high-quality solutions.
For our first deployment of this approach, within a single experiment we were able to find six hockey scoring exploit strategies that were qualitatively similar to those that playtesters had found in hours-long manual testing sessions.
\end{abstract}

\begin{IEEEkeywords}
Automated playtesting, Reinforcement Learning, Diversity
\end{IEEEkeywords}

\section*{Supplementary Video}
This paper is accompanied by videos of the agents trained with our algorithm, \algName: \url{go.ea.com/RAID}

\section{Introduction}
\begin{figure}[!htp]
    \centering
    \includegraphics[width=\linewidth]{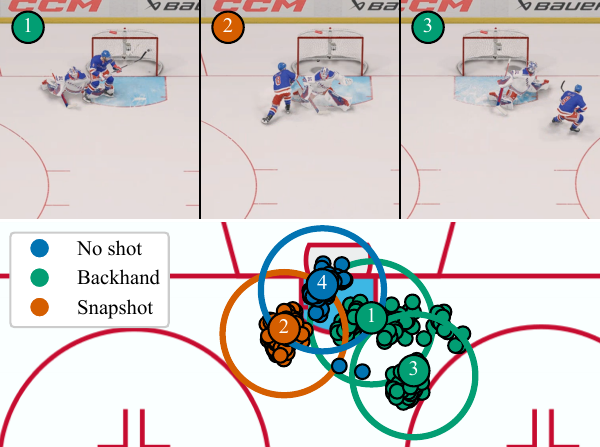}
    \caption{\textbf{Top: Each image shows the last frame before the goal of a different scoring strategy learned by an agent trained with \algName.} Our agent's forward player wears a blue jersey, the goalie a white jersey. The strategies are learned in an iterative fashion, with a reward function enforcing each new strategy's shot position to be at least $2$ meters away from all previous strategies using the same shot type.
    The numbers indicate which iteration learned this specific strategy and the colors indicate the shot types used to score the goal.\newline
    \textbf{Bottom: The shot locations and types of the strategies found by the first 4 iterations of \algName.} The small dots represent the shot location and type of $100$ goals taken after convergence of each iteration. The large circles represent a $2$ meter radius around the average position of those goals inside which no reward is given for all succeeding iterations of \algName\ if scoring with the same shot type. The shots by the agent of iteration 3 are for example outside of the $2$ meter radius of iteration 1, since the two agents use the same shot type.}
     \label{fig:overview}
  \end{figure}

With game worlds becoming vaster and game systems becoming more complex, the effort to test games grows likewise.
To reduce repetitive aspects of game testing, prior work has investigated automating parts of it through the use of autonomous agents playing a game and reporting bugs~\cite{ariyurek2019automated, de2022automated, bergdahl2020augmenting}.
While Reinforcement Learning (RL) agents can find individual exploits within a game~\cite{emergentToolUse}, they tend to converge to one single ``best'' solution of solving a task.
In this paper, we use \textit{EA SPORTS NHL 26} (short NHL) as our test-case. In particular, we aim to test the goalie AI of the game, training a forward agent to find high chance scoring strategies, which could represent potential exploits.
Our results indicate that standard RL algorithms tend to collapse to a small set of high-reward behaviors, rather than exploring a larger, more diverse set.
Thus, exploit discovery becomes sequential: developers must fix one issue before retraining the agent to find others.
Since this slows down the exploit finding process, we aim to develop an algorithm that can find multiple potential exploits without human intervention.

Prior work on RL for game testing has proposed methods to increase coverage of a game environment~\cite{gordillo2021improving,sestini2022automated}.
However, the added bias towards exploration comes at the cost of a loss in optimality with respect to the original goal of the agent.
For our use case, we are explicitly looking for high-performing scoring strategies and we therefore require a method that achieves high performance with respect to the scoring chance, while also finding multiple diverse solutions without human intervention.

The field of quality diversity investigates how to find multiple diverse solution that are high-performing~\cite{zahavy2022discovering}. While they have reported impressive results, existing solutions are often brittle in complex domains, one reason being a moving diversity target learned in parallel to the policies~\cite{leon2024discovering}.
We aim for playtesters to independently use our tool and therefore want the tool to be as simple as possible to require little RL expertise.

Following these requirements, we propose Reward-Adaptive Iterative Discovery (\algName), an approach to find a set of high-quality, diverse solutions by iteratively training a population of agents.
Figure~\ref{fig:overview} shows a brief overview of the results of the approach.
We enforce diversity through not rewarding agents for strategies that are similar to those found by any previously trained agent.
By measuring diversity over sequentially trained agents, leaving the diversity measure fixed during an individual agent's training time, we increase robustness compared to previous work.
We define diversity in an intuitive, domain-specific way, such that the diversity measure can be interpreted and adjusted by non-RL practitioners, facilitating the use by playtesters to extend their toolset.
\section{Related Work}

We present related work in two research directions connected to our approach: (1) automated playtesting, specifically approaches based on reinforcement learning, and (2) the field of behavior diversity, which motivates our approach.

\subsection{Automated Playtesting}

Contrary to recent work that apply reinforcement learning to train agents to master a game and surpass human players \cite{vinyals2019grandmaster, wurman2022outracing}, research on automated playtesting is generally less concerned with maximizing the same success metrics as human players, and more focused on having agents play a game in various ways that reveal insights on potential issues in the game.
One direction for this is to maximize the coverage of states an agent visits in a game.
A notable example is the work by \citet{gordillo2021improving}, that uses a count-based exploration reward scheme to promote visitation of less seen states in a game map.
Another direction aims to train agents that play games with behavior similar to human players, for example to visualize the interactions a game designer can expect from users or to balance the effectiveness of different strategies within a game~\cite{8295256, de2022automated, devlin2021navigation}.
\citet{sestini2022automated} combine both directions by exploring strategies similar to those of human players with an additional intrinsic reward for exploring new states.

While prior work achieves good results in terms of game state coverage and player-like behavior, we aim to investigate exploits that are closely tied to high performance with respect to the original human success metric of a game, in our case scoring goals with a high percentage in a hockey game.
We therefore aim to develop an approach to increase the diversity in behaviors of the trained agents while preserving the performance with respect to the game's success metric as much as possible.

\subsection{Diversity}

The field of quality diversity aims to generate diverse populations of high-performing
solutions through evolutionary algorithms~\cite{10.1145/2001576.2001606, mouret2015illuminating, cully2015robots}.
More recent work brings the concept of diversity into the field of reinforcement learning through intrinsic rewards.
The work by \citet{gregor2016variational}, for example, introduces an intrinsic reward based on optimizing the mutual information between options -- i.e. closed-loop
policies for taking action over a period of time -- and the options' final states, using entropy optimization to maximize diversity across options and distinction of individual options.
Similarly \citet{eysenbach2018diversity} optimize diversity over all visited states and show that this form of unsupervised discovery can serve as an effective pretraining mechanism for reinforcement learning.
DOMiNO defines diversity in the space of state-action occupancy~\cite{zahavy2022discovering}, represented by successor features~\cite{barreto2017successor}. 
It combines the intrinsic diversity reward and an extrinsic reward specific to the environment with Lagrange multipliers to control the trade off between diversity and optimality with respect to the extrinsic reward.
While DOMiNO can train multiple diverse agents in parallel this way, \citet{leon2024discovering} highlight that the approach fails to learn diverse behaviors in complex domains and can suffer from instability due to the approach learning the successor feature representations in parallel to the agents' policies.

Due to this instability and our aim for a stable algorithm that can be used by non-RL practitioners, we train a group of agents in a simple, sequential fashion, by masking strategies similar to those found by a prior agent from the reward function of all subsequent agents.
The diversity measurement introduced by this method of reward masking is static for a specific agent within a single training iteration and not any more complex in practice than simple static reward shaping in standard reinforcement learning~\cite{ng1999policy}.
Similar to prior work~\cite{gregor2016variational}, we define diversity in terms of the final state of an agent.
Rather than maximizing the distance between states, we however introduce diversity via hard constraints on state difference, enforced through reward masking of prior behaviors.
This allows users of \algName~to easily update the algorithm to their desired level of diversity, for example in the later introduced diversity criterion for NHL by increasing the area used for the shot similarity criterion to a \SI{4}{\meter} radius.
In comparison, the diversity hyperparameters of prior work, for example relative to the maximum achievable reward in~\cite{zahavy2022discovering}, are significantly harder to interpret, especially for non-RL practitioners.
\section{Methodology}

Our goal is to create an RL-based approach to autonomously find multiple diverse, high chance scoring strategies for the NHL environment without any human intervention in between iterations.
We first outline the base RL setup to find a single exploit within the NHL environment.
We then describe \algName, a novel approach to satisfy the posed requirements.
For the foundations of reinforcement learning, such as the concepts of reward, policy, and Q-value function, we refer the reader to~\cite{Sutton1998}.

\subsection{Base RL Setup}\label{sec:base}

\minisection{Action Space.} We aim to train a hockey forward player agent that learns high chance scoring strategies in a one-versus-goalie setting, similar to those that a human player could find by exploiting weaknesses in the goalie AI system.
To make sure that the found strategies are executable by a human, we provide the agent with a selection of the same actions available to human players on a game controller:
\begin{itemize}
  \item Discrete actions: the $4$ face buttons and the left trigger button that execute so-called {\it dekes}, i.e., decoy movements to draw the goalie out of position.
  \item Continuous actions: the two sticks for player skating and hockey stick control, each represented by continuous $x$ and $y$ axes.
\end{itemize}
Through the use of those actions, the agent can perform different kinds of shot types -- backhand, wrist shot, etc. -- which we will later consider for defining shot diversity.
Furthermore, we only let the agent act every $5$ frames and smooth the stick actions using an exponential moving average with a smoothing factor of $0.2$, to make sure that the agent does not leverage any super-human movement capability or reaction time.

\minisection{Observation Space.} We provide the agent with the normalized position, velocity, and orientation of puck, net, and goalie relative to the agent, as well as a stack of the $8$ last performed actions.

\minisection{Reward Function.} We reward the agent for every goal. To simplify learning, we additionally include shaping rewards for the puck getting closer to the goal, as well as for the scoring chance of every shot taken, based on an NHL-internal calculation.

\minisection{Algorithm \& Architecture.} While from an interface perspective all RL algorithms that use a reward function are technically compatible with \algName, \algName's effectiveness depends on the base algorithm's ability to explore various strategies -- even when previously found strategies are no longer rewarded -- before converging to the best one. 
We therefore use Soft Actor-Critic~\cite{haarnoja2018soft}, since it features a robust exploration mechanism through entropy maximization, exploring more in less known states.
Furthermore, Soft Actor-Critic has a high sample efficiency, which is crucial for environments with slow data generation, such as full-scale games.
To support both discrete and continuous actions, we use the architecture described by~\citet{delalleau2019discrete} for $2$ Q-value functions, as well as a mixed-action policy, each with $5$ hidden layers with $512$ units, allowing for a maximum of one discrete action at a time.
When we use the term \textit{agent}, we refer to both the policy and Q-value functions used for training and inference.
For efficient training, we stop training once an agent either reaches a $90$\% scoring chance or does not improve its performance for $50$k training steps.

\subsection{\algNameLong}

\begin{algorithm}[!t]
\caption{\algNameLong}\label{alg:RAID}
\begin{algorithmic}

    \State \textbf{Input:} $N$, $Z_\text{prev}$
    \State \Comment{$Z_\text{prev}$: previous strategies, can be $\emptyset$ or warm-started}
    \State \textbf{Output:} $Z_\text{prev}$
    \For{$i = 1$ to $N$}
        \State $r_i$($z$) $\gets$ $r_\text{base}$($z$) IF !\texttt{similar}($z$, $Z_\text{prev}$) ELSE $0$
        \State \Comment{$z$: strategy, $r$($z$): reward function}
        \State Initialize agent $\pi_i$
        \While{NOT \texttt{converged}($\pi_i$)}
            \State \texttt{train}($\pi_i$, $r_i$)
        \EndWhile
        \State $z_i \gets$ \texttt{evaluate}($\pi_i$)
        \State $Z_\text{prev} = Z_\text{prev} \cup \{z_i\}$
    \EndFor
\end{algorithmic}
\end{algorithm}

We outline the general, domain-agnostic \algName~method in Algorithm~\ref{alg:RAID}.
The algorithm extends a standard reinforcement learning setup by training multiple agents sequentially, aiming for each agent to come up with a new strategy for maximizing the reward of the environment.
For NHL we define a strategy $z$ as the shot type and shot position at the end of an episode, the shot position meaning the last position of the puck before the puck is shot or, if the puck is not shot, the last puck position before a goal. 
We start by training a first agent with a standard reinforcement learning setup.
After training of the first agent converges, we evaluate the agent to capture the strategy of its highest performing checkpoint.
For the NHL game, we evaluate the agent until it scores $100$ goals and describe the strategy $z$ as the average shot position and the most common shot type over those $100$ goals.
We store this strategy in a list of previous strategies $Z_\text{prev}$.
We then continue sequentially, training each next agent the same way, but for each new iteration modify the reward function to not reward strategies similar to those found by any previous agent.
Based on feedback from the NHL development team, we define \texttt{similar}($z$, $Z_\text{prev}$) as the shot position of a strategy being within a \SI{2}{\meter} radius of a previous strategy that used the same shot type, i.e. a strategy $z$ is considered novel or sufficiently diverse with respect to a set of existing strategies $Z_\text{prev}$ if: 
\[
\forall z' \in Z_\text{prev}\quad z_\text{shottype} \neq z'_\text{shottype} \vee l^2(z_\text{shotpos},z'_\text{shotpos}) >\text{\SI{2}{\meter}}.    
\]

For NHL, we skip adding strategies with a scoring chance of less than $10$\% after convergence, since we are interested in high-performing strategies and convergence at this low performance potentially indicates that the algorithm got stuck in a local optimum, meaning that future iterations starting with a different random seed could potentially find more performant solutions for the same shot position and type.
For NHL, we use the details from Section~\ref{sec:base} for the base reward function, training algorithm, and convergence criterion.

While the definition of strategy similarity is highly domain specific, we argue that this is exactly what makes the method suitable for game development.
While previous work defines diversity in a more general way that can be used across domains, for example via successor features~\cite{zahavy2022discovering}, we argue this makes it significantly harder for a playtester without RL knowledge to tune diversity to a level fitting their domain, given that~\cite{zahavy2022discovering}, for example, requires the diversity level to be expressed with respect to the maximum achievable reward.
Furthermore, having an explicit list of previous strategies allows playtesters to warm start training using a list of manually defined strategies, instead of an empty set, for example using high scoring strategies that are known and desired by the game, or exploits that are known but the game studio could not fix yet.

\section{Experiments}

We compare \algName~to a naive baseline to demonstrate the diversity in strategies found by the approach.
This is most notably supported by \algName~finding $6$ exploit strategies that were previously also found by human playtesters, but had not been fixed on the outdated version of the game that we conducted the experiments on.
We furthermore test the limitations of the algorithm with respect to the number of high chance scoring strategies it finds before failing to find any new strategies.

\subsection{Experimental Setup}
We run all experiments on a single PC with an \textit{NVIDIA GeForce RTX 4090} GPU and an \textit{AMD Ryzen Threadripper PRO 7975WX} CPU.
We use a pre-release development version of \textit{EA SPORTS NHL 26}, additionally reducing rendering to a minimum to speed up the simulation.
The simulation includes a single goalie, controlled by the game's AI we want to test, as well as a forward player, controlled by RL.
For both training and evaluation we slightly vary the initial position of the forward around the centerline to make the approach more robust to variations in the setup and to help with exploration.

\begin{figure}[!t]
 \centering
 \includegraphics[width=\linewidth]{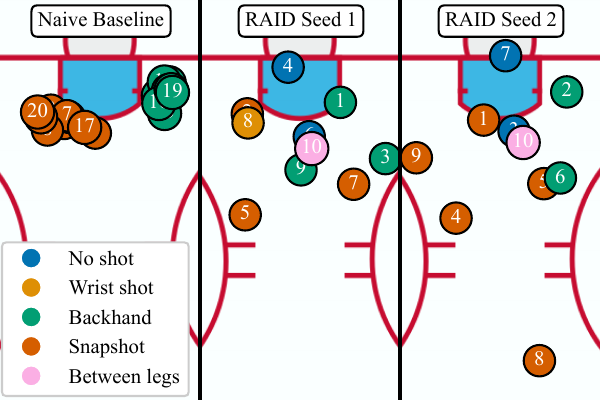}
 \caption{The average shot position and most common shot type after convergence for $20$ independent repetitions of the baseline experiment, as well as two experiments using \algName~started with different random seeds ran for $10$ iterations each. The color represents the most common shot type of an agent and the number represents the iteration of the algorithm that leads to the shot strategy. The naive baseline shows that a standard reinforcement learning algorithm keeps finding the same small set of solutions when repeated multiple times, with the algorithm finding 9 snapshot strategies and 11 backhand strategies, all from similar positions. 
 The two experiments using \algName~not only find the two strategies found by the baseline, but additionally also find 8 more strategies that adhere to our definition of diversity, maintaining a distance of at least \SI{2}{\meter} between shot strategies that use the same shot type.
}
 \label{fig:comparison}
\end{figure}
 
\subsection{Naive Baseline}\label{sec:exp_naive}
As a baseline, we run the base RL setup from Section~\ref{sec:base} for $20$ independent iterations, each with a different random seed, changing the initialization of the policy and value networks, as well as the random states of the game environment.
Each iteration uses the full reward function, assigning rewards for all goals, independent of shot position or type, without masking previous strategies.
The experiments take between \SI{16}{\min} and \SI{152}{\min} to converge.
RL algorithms aim to find the best solution to a problem with respect to a reward function.
In environments with a set of clearly superior strategies, returning a higher reward, an optimal algorithm therefore always converges to this same set of solutions, if not otherwise incentivized.
The left of Figure~\ref{fig:comparison} shows how all of the $20$ independent iterations end up with either one of two patterns, a snapshot from the left side of the goal or a backhand shot from the right side.

Assuming that a playtester classifies the found strategies as unwanted exploits, they have to improve the goalie AI behavior to mitigate those exploits before the baseline approach is likely to find any additional ones.
The naive RL approach therefore adds considerable overhead to the exploit finding process by requiring human intervention between iterations.
If on the contrary the playtester considers the found strategies to be a desired way of scoring in the game, they would need to find a way for the reward function to distinguish between exploits and desired scoring strategies, such that the agent ignores the desired strategies and finds novel exploits.
Exploits are highly unique in their appearance and therefore hard to define before first seeing them.

\subsection{\algNameLong}

We run \algName~starting with two different initial random seeds, influencing the random initialization of the agents of all iterations, for $N=30$ and $N=10$ iterations respectively.
The $30$ iteration experiment takes \SI{48}{\hour} and the $10$ iteration experiment takes \SI{14}{\hour} for all iterations to converge.
Figure~\ref{fig:comparison} shows how the strategies found by \algName\ adhere to the diversity criteria we set for NHL: while the average shot positions of strategies with the same shot type are all at least \SI{2}{\meter} away from each other, for example the three backhand shot strategies in green for seed $1$, the algorithm still finds strategies with close shot position but different shot types, e.g. strategies $8$ and $2$ for seed $1$, sharing almost the same position.
This way the algorithm finds a set of diverse strategies without any human intervention in between iterations to fix prior exploits, as would be required for the naive baseline.

As an illustrative example of the usefulness of \algName~for playtesting, when we ran \algName\ for the first time and presented the results to the development team, $6$ of the found strategies matched with exploits previously found by human playtesters.
Since we did however use an outdated version of the game for our experiments, those exploits were still available for the agents to find.
This shows that the agents are able to find exploits qualitatively similar to those found by human playtesters, while only requiring human supervision after the algorithm has found a set of candidates for a human expert to review. 
To not reveal any information that allows players to further exploit game mechanisms, we do not reveal which of the found strategies match with those deemed exploits.

\begin{figure}[!t]
 \centering
 \includegraphics[width=\linewidth]{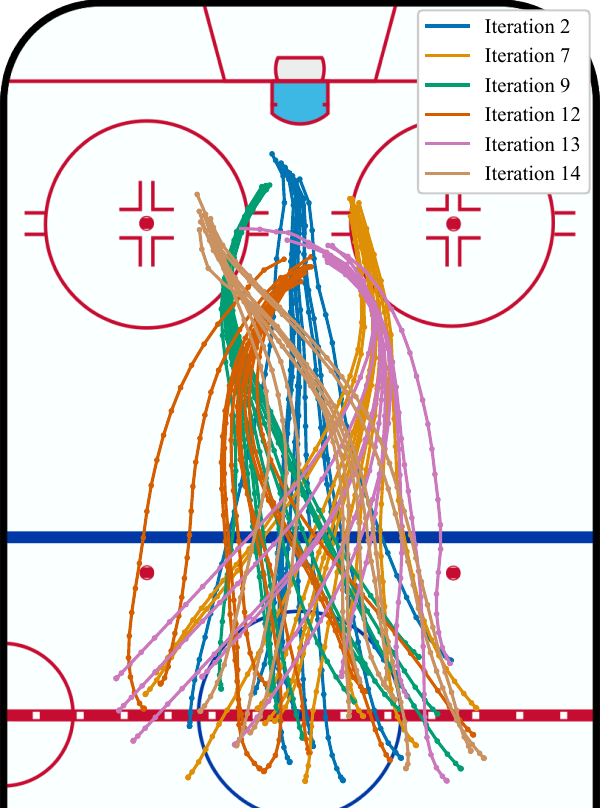}
 \caption{Movement of the agents of a hand-picked selection of iterations of seed $1$ of \algName~for $10$ different random starting positions each. This demonstrates that while diversity is only defined in terms of shot type and position of the shot, the positions before the shot, are also implicitly impacted for the agent to create a high scoring chance. For example, while iterations $12$ and $13$ have almost the same shot position, they approach the position from the opposite side to facilitate different shot types.}
 \label{fig:trajectories}
\end{figure}

\minisection{Pre-shot Diversity.} Figure~\ref{fig:trajectories} shows $10$ trajectories each for a selection of iterations of seed $1$ of \algName.
While we only define diversity in terms of shot position and type, the visualized trajectories demonstrate how this diversity definition implicitly generates diversity in states before the shot as well.
This shows that the agents take into account player direction, as well as orientation and stick position, which are visible in the supplementary videos, when learning scoring strategies.
While trajectories found by iterations $12$ and $13$, for example, have similar shot positions, they require significantly different skating trajectories to score with their respective shot types, backhand and snapshot.

\minisection{No-shot Goals.} To our surprise, the agent also finds strategies to score without shooting, marked in dark blue in Figure~\ref{fig:comparison} and visible in the supplementary video.
This initially led to a bug in our setup, since previous shot strategies were only recorded when the agent actually shot the puck.
This further underlines the difficulty of defining exploits before first seeing them, outlined in Section~\ref{sec:exp_naive}.

\minisection{Re-testing after Behavior Update.} While most of the strategies found in seed $1$ and seed $2$ resemble each other, they appear in different order and do not match completely -- for instance, strategies found in iteration $8$ of seed $1$ and iteration $8$ of seed $2$ are not found by the other seed.
RL algorithms have a high variance in their exploration process and therefore converge to different solutions given the existence of multiple similarly performant solutions, as we also demonstrate with the two dominant solutions the naive baseline finds.
This however means that if a playtester attempts to fix an exploit found by a first run of \algName, the same exploit not appearing in a second run of \algName~after the fix is not a sufficient guarantee that the fix was successful.
Similarly, the policies trained in the first run can not be used for this purpose, since RL is highly overfitting to the exact dynamics of the environment and can therefore be easily thrown off by slight changes in the goalie's behavior, without proving that the exploit is fixed.
Playtesters therefore still have to validate that an exploit is indeed fixed by imitating the strategy found by the approach before the fix.

\begin{figure}[!t]
 \centering
 \includegraphics[width=\linewidth]{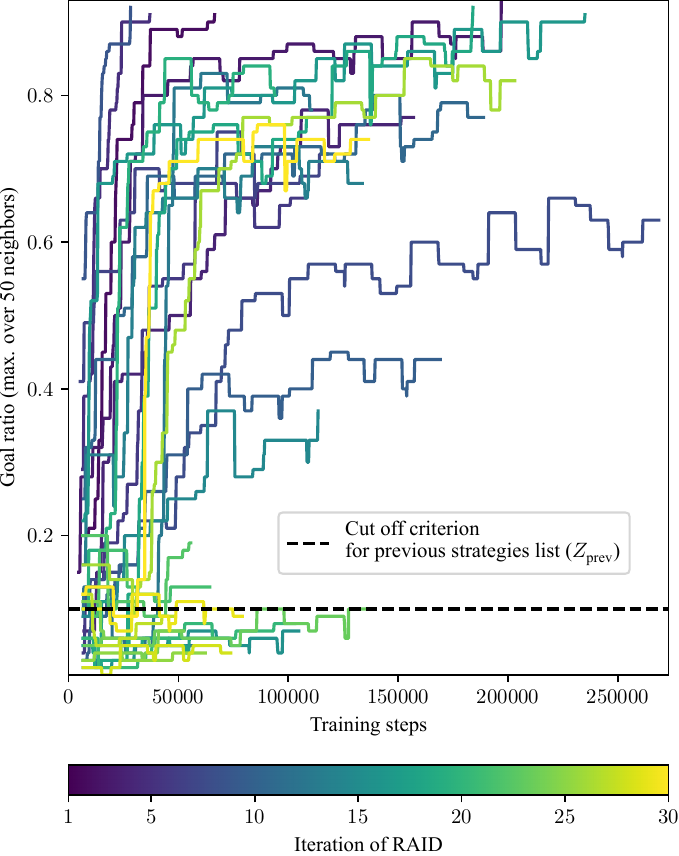}
 \caption{Training progress of seed $1$ of \algName\ running for $30$ iterations, represented by the goal rate of the agents during training episodes. For this figure, we smooth the goal rate using a maximum over $50$ neighbors to maintain information on whether an iteration reaches the goal rate threshold of $10$\% at any point in training. The coloring shows that later iterations of \algName, which are represented in brighter colors, are less likely to discover high chance scoring strategies, because most of those strategies were already found by previous iterations, represented in darker colors.}
 \label{fig:training_curves}
\end{figure}

\minisection{Iteration Limit.} Figure~\ref{fig:training_curves} shows the training progress of the $30$ iterations of seed $1$.
The first $10$ iterations all reach performances above the $10$\% goal rate cut-off criterion before triggering any of the early stopping criteria.
The longer the experiment goes on, the more iterations do not make the cut, with iterations $11$, $15$, $17$, $20$, $21$, $23-25$ and $27-29$ all converging at goal rates below the threshold and therefore not being added to the previous strategies list ($Z_\text{prev}$).
We manually stop the search after $30$ iterations, since we only find $3$ valid strategies in the last $11$ iterations, reducing the likelihood of discovering further interesting strategies in practical time.
More generally, the iteration limit allows for a trade-off between run time and the expected number of diverse solutions found, with the optimal choice being highly domain-specific and dependent on the diversity criterion.
In NHL for example, if we did not differentiate the shot type and only used the position as discriminative feature, we would expect to find less solutions over all, reducing the optimal search time.

\begin{figure}[!t]
 \centering
 \includegraphics[width=\linewidth]{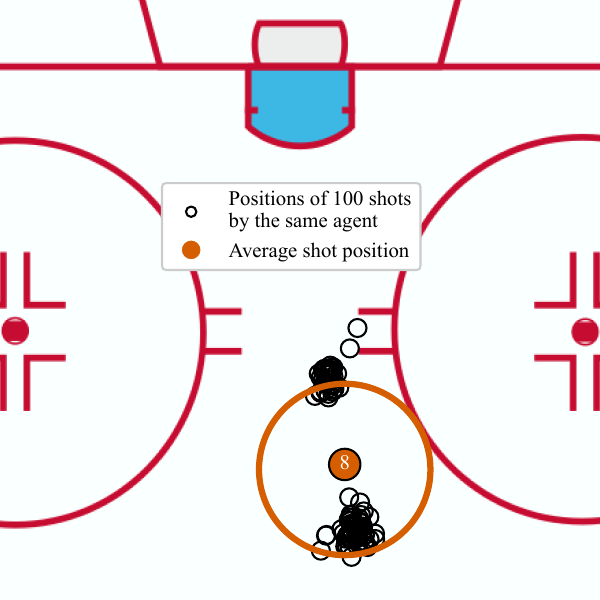}
 \caption{$100$ shot positions of the agent in iteration $8$ of seed $2$ of \algName, as well as the average of those shot positions.
 In succeeding iterations, shots within \SI{2}{\meter} of that average position are no longer rewarded if they use the same shot type as the most common shot type among those $100$ shots.
The agent learns a bi-modal behavior, shooting from either one of two shot positions depending on its spawn position. This leads to around $1/3$ of shots not being within the radius.
 Succeeding iterations of the algorithm are therefore still rewarded for shots similar to this part of the behavior, meaning that the exclusion mechanism based on the radius around the average does not completely exclude this scoring strategy. 
All shots are of the same shot type, wrist shot, though we have previously even seen bi-modal behavior with differing shot types}
 \label{fig:bimodal}
\end{figure}
\minisection{Multimodality.} Figure~\ref{fig:bimodal} shows the shot position of $100$ goals after training converged for the $8$th iteration of seed~$2$.
Around one third of all shot positions lie outside of the \SI{2}{\meter} radius around the average of all shot positions, meaning that in succeeding training iterations of \algName~the exact same shots will still be rewarded.
This highlights a potentially bigger problem around the introduced diversity criterion for NHL: in a more extreme hypothetical case, a single agent can learn to shoot from two positions that are more than $4$ meters apart, going for either of the two shot positions at a rate of $50$\% depending on where the agent is spawned.
This way, most of the shots of the agent could be outside of the \SI{2}{\meter} radius around the average of the shots' positions, meaning that future iterations could still learn the exact same behavior without it being excluded.
This motivates further investigations into more elaborate representations of a policy's strategy, for example through clustering methods such as k-means~\cite{mcqueen1967some}.

\section{Conclusion, Limitations, and Future Work}

We introduce \algNameLong, a method to find multiple diverse high chance scoring strategies without human intervention in between iterations.
The algorithm extends a standard reinforcement learning setup by training multiple agents sequentially, aiming for each agent to come up with a new strategy by not rewarding agents for strategies similar to those of previous iterations.
As a domain expert's review of the learned strategies shows, \algName~finds exploits similar to those identified by human playtesters.
We furthermore argue that defining diversity with respect to NHL allows game testers to better follow and adjust the method, compared to prior work that requires RL expertise to interpret the hyperparameters for tuning diversity.

\minisection{Limitations.} While our method finds potential exploits autonomously, it still requires human supervision post training, since telling high chance scoring strategies intended by game designers and actual exploits apart is not covered by the method.
Furthermore, due to the high variance in RL's exploration process, our method is not sufficient to determine whether a found exploit has been fixed successfully by an engineer, relying on playtesters for this verification task.

Our method's runtime grows (at least) linearly with the number of strategies we aim to find.
While previous work trains multiple diverse behaviors in parallel~\cite{zahavy2022discovering}, we deliberately forego this option to keep our method simple and therefore facilitate its use by playtesters not familiar with RL.

\minisection{Future Work.}
We design parts of the \algName~method specifically for the use on NHL, though the base idea outlined in Algorithm~\ref{alg:RAID} could be applicable to other domains as well.
While diversity is somewhat intuitive to define for the domain of scoring goals in NHL, future work could research how diversity can be defined in more complex domains, including multiplayer scenarios or even agents outputting text as actions.
While previous work that defines diversity in a domain-agnostic way often requires RL knowledge to interpret and adjust hyperparameters, it would be interesting to research definitions of diversity that are domain-agnostic but still interpretable by non-RL practitioners, for example by defining the minimum distance of trajectories used in \algName~in a more general way.

\bibliographystyle{IEEEtranN}
\bibliography{ref}

\end{document}